\documentclass[11pt,a4paper]{article}
\usepackage[colorlinks=true,linkcolor=blue]{hyperref}

\usepackage{cmap}
\usepackage[utf8x]{inputenc}
\usepackage[T1]{fontenc}
\usepackage[english]{babel}
\usepackage{amssymb,amsmath}
\usepackage{amsthm}
\usepackage{bbm}
\usepackage{microtype}
\usepackage{lmodern}

{
\rmfamily
\DeclareFontShape{T1}{lmr}{m}{sc}{<->ssub*cmr/m/sc}{}
\DeclareFontShape{T1}{lmr}{b}{sc}{<->ssub*cmr/b/sc}{}
\DeclareFontShape{T1}{lmr}{bx}{sc}{<->ssub*cmr/bx/sc}{}
}

\newcommand{\thmheadercommand}[1]{\textbf{\scshape{}#1.\\*}}

\newtheoremstyle{yannthm}{\topsep}{\topsep}{\slshape}{}{\scshape\bfseries}{.}{.5em}{%
\thmname{#1}\thmnumber{ #2}\thmnote{#3}%
}
\newtheoremstyle{yannthm2}{\topsep}{\topsep}{}{}{\scshape\bfseries}{.}{.5em}{%
\thmname{#1}\thmnumber{ #2}\thmnote{#3}%
}

\def\d{\operatorname{d}\!{}}

\def\R{{\mathbb{R}}}

\renewcommand{\leq}{\leqslant}

\newcommand{\deq}{\mathrel{\mathop{:}}=}

\def\eps{\varepsilon}
\renewcommand{\epsilon}{\varepsilon}
\renewcommand{\phi}{\varphi}

\let\oldPr\Pr
\renewcommand{\Pr}{\oldPr\nolimits}
\newcommand{\E}{\mathbb{E}}

\DeclareMathOperator{\Id}{Id}

\newcommand{\norm}[1]{\left\lVert#1\right\rVert}

\newenvironment{dem}[1][]{\begin{proof}[\thmheadercommand{Proof#1}]~\newline\ignorespaces}{\end{proof}}

{
\theoremstyle{yannthm}
\newtheorem{defi}{Definition}
\newtheorem*{defi*}{Definition}
\newtheorem{prop}[defi]{Proposition}
\newtheorem*{prop*}{Proposition}
\newtheorem{thm}[defi]{Theorem}
\newtheorem*{thm*}{Theorem}

\newtheorem*{lem*}{Lemma}
\newtheorem{cor}[defi]{Corollary}
\newtheorem*{cor*}{Corollary}

\newtheorem*{ex*}{Example}

\newtheorem*{subenonce}{}

\theoremstyle{yannthm2}

\newtheorem*{exo*}{Exercise}

\newtheorem*{rem*}{Remark}

\newtheorem*{subenonce2}{}

}

\newcommand{\transnorm}[1]{\norm{#1}_\mathrm{Dir}}
\newcommand{\Penv}{P_\mathrm{env}}

\newcommand{\munorm}[1]{\norm{#1}_\mu}
\newcommand{\scalmu}[2]{\langle #1,#2\rangle_\mu}

\newcommand{\pp}{\phi}
\newcommand{\AR}{\mathcal{R}}

\title{Approximate Temporal Difference Learning is a Gradient Descent for
Reversible Policies}
\author{Yann Ollivier}
\date{}

\begin{document}

\maketitle

\begin{abstract}
In reinforcement learning, temporal difference (TD) is the most direct
algorithm to learn the value function of a policy. For large or infinite
state spaces, exact representations of the value function are usually not
available, and it must be approximated by a function in some parametric
family.

However, with \emph{nonlinear} parametric approximations (such as neural
networks), TD is not guaranteed to converge to a good approximation of
the true value function within the family, and is known to diverge even
in relatively simple cases.  TD lacks an interpretation as a stochastic
gradient descent of an error between the true and approximate value
functions, which would provide such guarantees.

We prove that approximate TD is a gradient descent provided the current
policy is \emph{reversible}. This holds even with nonlinear
approximations.

A policy with transition probabilities $P(s,s')$
between states is reversible if there exists a function $\mu$ over states
such that $\frac{P(s,s')}{P(s',s)}=\frac{\mu(s')}{\mu(s)}$. In
particular, every move can be undone with some probability. This
condition is
restrictive; it is satisfied, for instance, for a navigation
problem in any unoriented graph.

In this case, approximate TD is exactly a gradient descent of the \emph{Dirichlet
norm}, the norm of the
difference of \emph{gradients} between the true and approximate value
functions. The
Dirichlet norm also controls the bias of approximate policy gradient.
These results hold even with no decay factor ($\gamma=1$) and do not
rely on contractivity of the Bellman operator, thus
proving stability of TD even with $\gamma=1$ for reversible policies. 
\end{abstract}

The temporal difference (TD) algorithm is a cornerstone of reinforcement
learning, allowing for computation of the Bellman value function of a
given policy \cite{suttonbarto98}. However, with large or continuous search spaces,
maintaining the exact value function at each state is unfeasible, and
parametric approximations of the value function are used instead
\cite[\S 8]{suttonbarto98}.

With such parametric approximations, TD is not guaranteed to
converge to the best approximation of the true value function within the
family, or even, to converge at all \cite[\S X]{tsitsiklis1997approxtd}. This is in great part because the TD
algorithm lacks an interpretation as a stochastic gradient descent of an
error between the true and approximate value functions.

For \emph{linear} families of approximating functions, TD is known to
converge to some fixed point \cite{tsitsiklis1997approxtd}; this fixed
point is related, but generally not identical, to the best approximation
in the family.  For nonlinear approximations, TD is known to diverge even
in relatively simple cases. Current popular families using neural
networks are nonlinear.

% We claim that, on-policy and \emph{provided the
% Markov chain of the policy is reversible}, approximate TD learning
% converges to a locally best function in the class that approximates the true
% $V$ function in a weighted norm; the norm is given by the transition
% probabilities of the current policy, in a simple way. The class of
% approximating functions can be anything and is not restricted to affine
% families.

% The reversibility assumption on the policies basically restricts the
% theorem to navigation-type problems, in which the agent directly selects
% the next state among a set of neighbors, and any move can be reversed.

As a theoretical study of nonlinear value function approximation,
\cite{nonlinearprojtd2009}
introduces an algorithm more complex than TD, involving
second derivatives of the approximating family. This algorithm has an
interpretation as a gradient descent of an objective function $J$. $J$ is
built so that the global minimum of $J$ is also a fixed point of TD;
however, the algorithm may also converge to a local minimum of $J$ with
unclear significance. Moreover this does not address the interpretation
of fixed points of TD in the first place.

Here we consider the unmodified approximate TD algorithm, with any
class of approximating functions, linear or not. We
prove that approximate TD coincides with a gradient descent of the \emph{Dirichlet
norm} of the error between the true and approximate value functions
(Theorem~\ref{thm:revTD}), provided the current policy is
\emph{reversible}.

Reversibility (see Section~\ref{sec:notation}) is a common assumption in
the mathematical treatment of Markov chains, because of its convenience.
It implies that any allowed transition between states can also occur in
reverse with some probability. It is satisfied, for instance, by the
random walk on unoriented graphs, or by Brownian motion and other
stochastic processes.

The Dirichlet norm is used in the treatment of the
convergence of Markov chains \cite{DSC_logsob,LPW2009_markov}, and is directly related to the spectral gap
of the random walk operator. This norm is given in a simple way by the
transition probabilities of the current policy (Eq.~\ref{eq:dir}). Its natural appearance in approximate TD is
perhaps remarkable.

Therefore, approximate TD learning will minimize the approximation error in
Dirichlet norm, for reversible policies. Interestingly, this minimization
also directly controls the bias of approximate policy gradient, which
also involves the Dirichlet norm (Proposition~\ref{prop:dirgrad}).

However, in a reinforcement learning setting, the reversibility
assumption is quite restrictive. First, it implies that any move can be
undone with some probability. Second, reversibility depends both on the
policy and the environment (via Eq.~\ref{eq:transprob}); in general,
reversibility cannot be checked knowing the policy alone. An exception to this
are \emph{navigation-type} problems, in which the policy consists in
directly choosing the next state among a set of possible states (e.g.,
exploring an undirected graph).
For such problems, it is easy to check reversibility, and to keep the
policy reversible at all times, e.g., by using a Gibbs policy with respect to some
energy function on state space (see Section~\ref{sec:revpolicies}).

Thus, although we have stated each result under general mathematical
assumptions, the results here chiefly make sense in navigation-type problems, in
which the agent directly selects the next state among a set of neighbors,
and any move can be reversed. 

\paragraph{Acknowledgments.} I would like to thank Léon Bottou,
Alessandro Lazaric, Corentin Tallec and Nicolas Usunier for pointers to
references and for suggestions on the text.

\section{Notation and Markov Chain Background}
\label{sec:notation}

\paragraph{Markov decision processes.} 
We mostly borrow notation from \cite{nonlinearprojtd2009}.
Consider a finite\footnote{The arguments presented here do not
crucially rely on finiteness: algebraically the results would hold for a countable or
continuous state space as well, as long as all sums and expectations are
well-defined. We consider the finite case to avoid measurability issues.}
Markov decision process (MDP) and a policy $\pi$ for this MDP.
Let $\pi(s,a)$ be the probability to select action $a$ when in
state $s$. Let $\Penv((s,a),s')$ the probability that the environment
jumps to $s'$ after that.
Let $r_1,\ldots,r_t$ be the sequence of rewards of this MDP:
$r_t$ is the reward incurred while arriving in state $s_t$, a
random variable depending on $a_{t-1}$ and $s_{t-1}$.

Given an initial state $s_0$, denote
$\E_{\pi,s_0}$ the expectation under a random sequence
$(a_0,s_1,a_1,\ldots)$ of actions and states resulting from $\pi$ and
$\Penv$, defined inductively by $a_t\sim \pi(s_t,\cdot)$ and $s_{t+1}\sim
\Penv((s_t,a_t),\cdot)$.

% More generally, for a given finite sequence $s_0,a_0,s_1,\ldots$ we denote
% $\E_{\pi,s_0,a_0,s_1,\ldots}$ an expectation conditioned on this initial
% sequence of actions and states.

% Given a policy $\pi$, let $V^\pi$ be the expected
% rewards from that policy (Bellman value function) with decay factor
% $\gamma\leq 1$, defined as
% \begin{equation}
% V^\pi(s)\deq \E_{s,\pi} \sum_{t=1}^\infty \gamma^{t-1} r_t
% \end{equation}
% with $r_t$ the reward obtained at time $t$, which is a random function of
% $(s_t,a_{t-1})$.
% 
% We say that $(\pi,V^\pi)$ are a \emph{Bellman pair} if $V^\pi$ is the
% expected reward for $\pi$, and $\pi$ is the greedy policy based on
% $V^\pi$, namely,
% \begin{equation}
% \pi(s,a)>0 \Rightarrow a\in \argmax_a \left\{ \E_{s,a} \sum_{t=1}^\infty
% \gamma^{t-1}r_t\right\}=\argmax_a \{ \E_{s,a} [r_1+\gamma V^\pi(s_1)]\}
% \end{equation}
% This is the optimal expected reward under the given MDP.

% Let $Q(s,a)$ be a function of states and actions. We say that $Q$
% \emph{describes expected rewards from policy $\pi$} (is the $Q$-function for
% $\pi$) if
% \begin{equation}
% Q(s,a)=\E_{\pi,s,a} \sum_{t=1}^\infty \gamma^{t-1}r_t
% \end{equation}
% for all $s,a$. Likewise, we say that $V(s)$ \emph{describes rewards from
%$\pi$} or \emph{is the value function of $\pi$} if
The \emph{value function} of policy $\pi$ in state $s$, with \emph{decay parameter}
$\gamma<1$, is
\begin{equation}
V(s)\deq\sum_{t=1}^\infty \gamma^{t-1}\E_{\pi,s}[ r_t]
\end{equation}

% We say that $\pi$ is a \emph{greedy policy from $Q$} if $\pi$ selects the
% actions $a$ with largest value $Q(s,a)$ when in state $s$, namely, if
% \begin{equation}
% \pi(s,a)>0 \Rightarrow a \in \argmax_a Q(s,a)
% \end{equation}
% 
% We say that $(\pi,Q)$ is a \emph{Bellman pair} if, at the same time,
% $\pi$ is greedy from $Q$ and $Q$ describes expected rewards from $\pi$.
% By classical arguments, such a $\pi$ provides optimal expected reward under the given MDP.

% [DUPLICATE] More precisely, consider, as above, a finite MDP with policy $\pi(s,a)$ and
% transition probabilities $\Penv((s,a),s')$ from the environment.

Define
the transition probability matrix $P$ on states, that amounts to first selecting an
action according to $\pi$, then letting the environment select the next
state \cite{nonlinearprojtd2009}:
\begin{equation}
\label{eq:transprob}
P(s,s')\deq\sum_a \pi(s,a)\Penv((s,a),s')
\end{equation}

The value function for policy $\pi$ satisfies the Bellman
equation using transition probabilities $P$,
\begin{equation}
V=R+\gamma PV
\end{equation}
where $P$ and $V$ are seen as a matrix and vector, and
\begin{equation}
\label{eq:R}
R(s)\deq \E_{\pi,s} [r_1]
\end{equation}
is the average instantaneous reward of the policy in a given state.

For $\gamma=1$ the value function is usually infinite. We use the
\emph{relative} value function \cite[\S 6.7]{suttonbarto98}, also known
as \emph{bias} \cite[\S 5.1.1]{bertsekasvol2_ed4},
denoted $U$. Assuming that the current policy has a unique stationary distribution
$\mu$ over states, the relative value function is obtained by 
centering rewards:
\begin{equation}
U(s)\deq
\sum_{t=1}^\infty \E_{\pi,s} [r_t-\E_\mu R]
\end{equation}
where $\E_\mu R=\sum \mu(s) R(s)$ is the average reward under the
stationary distribution $\mu$. (Assuming ergodicity of $P$, this
expectation is finite in a finite MDP \cite[\S 5.1.1]{bertsekasvol2_ed4}, though without the
expectation the sum usually diverges as noise accumulates.) The relative value function satisfies the
Bellman equation with $\gamma=1$ and centered rewards \cite[Prop.~5.1.9]{bertsekasvol2_ed4}
\begin{equation}
\label{eq:cBell}
U=(R-\E_\mu R)+PU
\end{equation}

\paragraph{Approximate TD.} Let $V_\theta$ be an approximation to the
true function $V$, belonging to some family of functions smoothly parameterized by
$\theta$.

Given a transition $s\to s'$ with reward $r$, the gap in the Bellman
equation at $s$ is $r+\gamma V_\theta(s')-V_\theta(s)$. For the true $V$
function, this gap is $0$ on average (on average, because given $s$, the
state $s'$ and the reward are random). Approximate TD (e.g.
\cite[\S 8.2]{suttonbarto98} with $\lambda=0$) performs
an update
on $V_\theta(s)$ to reduce the gap,
\begin{equation}
\theta\gets \theta + \alpha \,\Delta \theta
\end{equation}
where $\alpha$ is a learning rate and $\Delta \theta$ is the update
\begin{equation}
\label{eq:approxTD}
\Delta\theta(s,s',r) \deq \left(r+\gamma
V_\theta(s')-V_\theta(s)\right)\partial_\theta
V_\theta(s)
\end{equation}

This gradient step has the effect of moving $V_\theta(s)$ closer to the
\emph{current} value of $r+\gamma V_\theta(s')$, ignoring the fact that
$V_\theta(s')$ will change as well.

\paragraph{Reversibility of Markov chains.} A Markov chain defined by the transition
matrix $P$ is \emph{reversible} \cite[\S1.6]{LPW2009_markov} if there exists a nonzero function $\mu$ on
states such that
\begin{equation}
\label{eq:rev}
\mu(s)P(s,s')=\mu(s')P(s',s)\quad \forall s,s'
\end{equation}
When nonzero this rewrites as $P(s,s')/P(s',s)=\mu(s')/\mu(s)$:
the ratio between the probability of a
transition and the reverse transition must be equal to a ratio of a
function of the target states.
In particular, any states $s$ and $s'$ with
nonzero $\mu$ must satisfy $P(s,s')>0\Leftrightarrow P(s',s)>0$.

For instance, the simple random walk in any
\emph{unoriented} graph is reversible with
$\mu(s)= \deg(s)$ \cite[\S1.6]{LPW2009_markov}.

When $P$ is reversible with respect to $\mu$, then $\mu$ (once rescaled)
is a stationary distribution of $P$ \cite[Prop.~1.19]{LPW2009_markov}. Indeed, the condition above
describes \emph{detailed balance}: if starting from distribution $\mu$, the
flow of mass from $s$ to $s'$ is equal to that from $s'$ to $s$, so that
every exchange is balanced and $\mu$ is stationary.

Therefore, reversibility of a Markov chain is usually expressed directly with respect to
its stationary distribution $\mu$.

On any unoriented graph, the Metropolis--Hastings construction provides
reversible random walks with arbitrary stationary distributions (see
Section~\ref{sec:revpolicies}).  Thus, for navigation problems on states
spaces with reversible moves, it would be easy to keep the policy
reversible.

By abuse of language, in a reinforcement learning context within a fixed
environment, we will call a policy \emph{reversible} if the Markov
chain $P$ defined by this policy in that environment via
\eqref{eq:transprob} is reversible.

\paragraph{The Dirichlet norm for Markov chains.}
Given a function $f$ on the state space, define its square norm under the
stationary distribution $\mu$, and the associated bilinear form, as
\begin{equation}
\munorm{f}^2\deq \sum_s \mu(s) f(s)^2
,\qquad
\scalmu{f}{g} \deq \sum_s \mu(s) f(s)g(s)
\end{equation}
The weighting by $\mu(s)$ is perhaps best interpreted as an average over
a long trajectory sampled from the policy.

The Markov chain is reversible with respect to $\mu$ if and only if $P$
%(equivalently $L$)
is self-adjoint for this bilinear form, namely, if and only if
$\scalmu{Pf}{g}=\scalmu{f}{Pg}$, where $P$ acts on a function $f$ over
states by
viewing $P$ as a matrix and $f$ as a vector. This is a direct consequence of
\eqref{eq:rev}.

We also define the \emph{Dirichlet norm} depending on the transition matrix $P$:
\begin{equation}
\label{eq:dir}
\transnorm{f}^2\deq \frac12 \sum_{s,s'} \mu(s) P(s,s') (f(s')-f(s))^2
\end{equation}
where $\mu$ is the invariant distribution on states resulting from $P$.
This quadratic form is actually a seminorm, since constant functions have norm $0$: adding a constant to $f$ does
not change $\transnorm{f}$. If $P$ is irreducible then constant functions are the
only such functions: if $\transnorm{f_1-f_2}=0$ then $f_1$ and $f_2$ are
equal up to an additive constant. This justifies the name \emph{norm}
if quotienting by constant functions.

$\transnorm{f}^2$ is often called the \emph{Dirichlet form} in the
Markov chain literature \cite{DSC_logsob,LPW2009_markov}. It
is a discrete Markov chain analogue of the
gradient norm $\int \norm{\nabla f}^2$ of a continuous function (the
classical ``Dirichlet form''): indeed, for $f$
a smooth function with compact support in $\R^d$, and $P$ the
nearest-neighbor random walk on
an $\eps$-grid in $\R^d$, with $\eps\ll 1$, at any point $x$ in the grid one has
\begin{equation}
\sum_{x'} P(x,x') (f(x')-f(x))^2=\frac{\eps^2}{d} \norm{\nabla f(x)}^2
+O(\eps^3)
\end{equation}
by a direct Taylor expansion, and therefore
\begin{equation}
\transnorm{f}^2=\frac{\eps^2}{2d}\int_{\R^d} \norm{\nabla f(x)}^2\d x +O(\eps^3)
\end{equation}

By elementary computations, the Dirichlet norm satisfies
\cite{DSC_logsob,LPW2009_markov}
\begin{equation}
\label{eq:lapnorm}
\transnorm{f}^2=\langle (\Id-P) f,f\rangle_\mu
\end{equation}
% where
% $L\deq \Id -P$ is the discrete Laplace operator of the random walk.

These two norms control one another up to centering: for any $f$,
\begin{equation}
\label{eq:eqnorms}
\beta \munorm{f-\E_\mu f}^2\leq \transnorm{f}^2\leq \munorm{f-\E_\mu
f}^2%\leq \frac{1}{\beta} \transnorm{f}^2
\leq \munorm{f}^2
\end{equation}
with $\beta$ the \emph{spectral gap} of the random walk
% (smallest non-zero
% eigenvalue of $\Id-P$)
\cite{DSC_logsob}. In practice $\beta$ may be quite small: e.g., for the
simple random walk on a cycle of length $n$, one has $\beta\approx
1/n^2$. Therefore $\transnorm{\cdot}$ can be significantly smaller than
$\munorm{\cdot}$.

\section{Approximate TD for Reversible Policies}

We claim that \emph{if $P$ is reversible with respect to its stationary
distribution $\mu$}, then approximate TD learning with a class of functions
$V_\theta$, tries to best approximate the true function $V$ by gradient
descent. The quality of the approximation is defined via
a mixed norm of $V_\theta-V$,
\begin{equation}
\gamma \transnorm{V_\theta-V}^2
+ (1-\gamma) \munorm{V_\theta-V}^2
\end{equation}
where $V$ is the true Bellman function associated with policy $P$.

For $\gamma$ close to $1$, the Dirichlet norm $\transnorm{V_\theta-V}$
dominates, while for small $\gamma$ the $\mu$-norm dominates. (For
$\gamma=0$
the $V$-function is equal to the
expected instantaneous reward.)

Thus, assuming reversibility, approximate TD will usually converge
to a local minimum of this mixed norm of $V_\theta-V$. This is independent
of the family of parametric approximations for $V$. For the particular
case of a linear family over $\theta$, the mixed norm is quadratic in
$\theta$, therefore convergence will be to a global minimum of the mixed
norm. The equivalence of the norms \eqref{eq:eqnorms} can be used to
transfer the minimization property to either $\munorm{\cdot}$ or
$\transnorm{\cdot}$ up to factors $\beta$.

\begin{thm}
\label{thm:revTD}
Consider a policy in some finite MDP.
Assume the policy is reversible, with stationary distribution $\mu$.

Let $V$ be the value function of the policy with decay factor $0\leq
\gamma< 1$. Let $(V_\theta(s))_\theta$ be a family of functions on the state
space, smoothly parameterized by $\theta$.

Then, on average over the stationary distribution $\mu$, the
step $\Delta\theta(s,s',r)$ made by approximate TD \eqref{eq:approxTD} is equal to a gradient
descent of a mixed norm of 
$V_\theta-V$,
\begin{equation}
\E_{s\sim \mu} \,\Delta\theta(s,s',r)= -\frac12 \partial_\theta \left(
\gamma \transnorm{V_\theta-V}^2
+ (1-\gamma) \munorm{V_\theta-V}^2
\right)
\end{equation}
where $s'$ and $r$ are the (random) next state and reward from state $s$.
\end{thm}

The theorem is in expectation over states $s$ from the stationary
distribution. Averaging over a
long enough trajectory, with small enough learning rates, will
approximate this expectation. \footnote{TD is a stochastic update
whose noise depends on $s$, so that the noise is Markov instead
of iid. The general theory of stochastic algorithms with Markov noise
from \cite{BMP90adapalgo} is used in
\cite{tsitsiklis1997approxtd} to offer a full treatment of TD for linear
approximations $V_\theta$.}

At the core of the proof, TD only takes into account cross-terms between
$\partial_\theta V_\theta$ at the current state and the value
function at the next state, while
the gradient of the error between $V_\theta$
and $V$ also comprises cross-terms between $\partial_\theta V_\theta$ at
the next state and the value function at the current state. In the
reversible case, the statistics of transitions $s\to s'$ and $s'\to s$
are identical in the stationary regime, hence TD is indeed a gradient of
the error.

\begin{dem}
The expected TD step in the stationary regime is
\begin{align}
\E_{s\sim \mu}\,\Delta\theta(s,s',r)&=\sum_s \mu(s) \partial_\theta
V_\theta(s)\,\E_{s'|s}\left[
r+\gamma V_\theta(s')-V_\theta(s)
\right]
\\&=\sum_s \mu(s) \partial_\theta
V_\theta(s)\left(R+\gamma (P V_\theta)(s)-V_\theta(s)\right)
\\&=\scalmu{\partial_\theta V_\theta}{R+\gamma PV_\theta-V_\theta}
\label{eq:avtd}
\end{align}
namely, the expected TD step is the dot product between the Bellman gap
$V_\theta$, and the direction of change
$\partial_\theta
V_\theta$ that can be realized within the parametric family. (In the linear case, this reduces to, e.g., Lemma 8 in
\cite{tsitsiklis1997approxtd}, with $\partial_\theta V_\theta=\Phi$.)

Define the difference between the approximated and true $V$ functions:
\begin{equation}
f_\theta\deq V_\theta-V
\end{equation}
we want to prove that the expected TD step is the gradient of the mixed
norm of $f_\theta$.

Since $V$ satisfies the Bellman equation $R+\gamma PV-V=0$ one has
\begin{equation}
R+\gamma PV_\theta-V_\theta=\gamma P f_\theta-f_\theta
\end{equation}
and moreover $\partial_\theta V_\theta=\partial_\theta f_\theta$ as $V$
does not depend on $\theta$. Therefore, \eqref{eq:avtd} rewrites as
\begin{align}
\E_{s\sim \mu}\,\Delta\theta(s,s',r)&=\scalmu{\partial_\theta
f_\theta}{(\gamma P -\Id)f_\theta}
\\&=
\label{eq:tddecomp}
%\E_{s\sim \mu}\,\Delta\theta(s,s',r) &=
-\gamma \,\scalmu{\partial_\theta
f_\theta}{(\Id -P)f_\theta}-(1-\gamma) \scalmu{\partial_\theta
f_\theta}{f_\theta}
\end{align}

Now the last term is the gradient of the $\mu$-norm:
\begin{equation}
\label{eq:gradmunorm}
\scalmu{\partial_\theta
f_\theta}{f_\theta}=\frac12 \partial_\theta \scalmu{f_\theta}{f_\theta}
=\frac12\partial_\theta \norm{f_\theta}^2_\mu
\end{equation}

Likewise, the first term is related to the norm $\transnorm{\cdot}^2$
thanks to \eqref{eq:lapnorm}:
\begin{equation}
\transnorm{f_\theta}^2=\scalmu{f_\theta}{(\Id-P)f_\theta}
\end{equation}
hence
\begin{equation}
\partial_\theta \transnorm{f_\theta}^2=\scalmu{\partial_\theta
f_\theta}{(\Id-P)f_\theta}+\scalmu{f_\theta}{(\Id-P)\partial_\theta
f_{\theta}}
\end{equation}
as $\Id-P$ is a linear operator that does not depend on $\theta$.

But the policy is reversible with respect to $\mu$ if and only if $P$ is
self-adjoint with respect to $\scalmu{\cdot}{\cdot}$. In that case,
\begin{equation}
\scalmu{f_\theta}{(\Id-P)\partial_\theta
f_{\theta}}=\scalmu{(\Id-P)f_\theta}{\partial_\theta f_\theta}
\end{equation}
and therefore
\begin{equation}
\label{eq:graddirnorm}
\partial_\theta \transnorm{f_\theta}^2=2\scalmu{\partial_\theta
f_\theta}{(\Id-P)f_\theta}
\end{equation}

Collecting \eqref{eq:gradmunorm} and \eqref{eq:graddirnorm} into
\eqref{eq:tddecomp}, we find
\begin{equation}
\E_{s\sim \mu}\,\Delta\theta(s,s',r)=-\frac12
\gamma\, \partial_\theta \transnorm{f_\theta}^2-\frac12 (1-\gamma) \,\partial_\theta
\norm{f_\theta}^2_\mu
\end{equation}
as needed.
\end{dem}

In the general, non-reversible case, the gradient descent of $\transnorm{V_\theta-V}^2$ differs from TD by
\begin{equation}
2\sum_{s,s'} \mu(s) P(s,s')
\partial V_\theta(s')
\left(
V_\theta(s')-V_\theta(s)-V(s')+V(s)
\right)
\end{equation}
which we cannot compute without knowing $V$. At least we would have to
know how to estimate $
\E_{s|s'} V(s)-V(s')$
given $s'$. That is, we would need to be able to sample backward transitions
leading to $s'$, and to evaluate the reward along these transitions. This
is similar to attempting to take the gradient of the squared Bellman
error \cite[\S 8.5]{suttonbarto98}.

We now turn to the case $\gamma=1$. The relative value function $U$ can
be approximated by using approximate TD with \emph{centered} rewards,
namely, by removing the stationary expected reward at each step \cite[\S
6.7]{suttonbarto98}.
In practice the expected reward is usually unknown and must be
approximated by averaging over the past.

\begin{thm}
\label{thm:revrelTD}
Consider a policy in some finite MDP.
Assume the policy is reversible, with stationary distribution $\mu$.

Let $U$ be the relative value function of the policy (with decay factor
$\gamma=1$).
Let $(U_\theta(s))_\theta$ be a family of functions on the state
space, smoothly parameterized by $\theta$. Let $\Delta\theta(s,s',r)$ be
the step made by centered approximate TD during a transition $s\to s'$ with
reward $r$, namely
\begin{equation}
\label{eq:relapproxTD}
\Delta\theta(s,s',r) \deq \left(r-\E_{\mu}R+
U_\theta(s')-U_\theta(s)\right)\partial_\theta
U_\theta(s)
\end{equation}

Then, on average over the stationary distribution $\mu$, the
step made by centered approximate TD is equal to a gradient
descent of the Dirichlet norm of
$U_\theta-U$,
\begin{equation}
\E_{s\sim \mu} \,\Delta\theta(s,s',r)= -\frac12 \,\partial_\theta
\transnorm{U_\theta-U}^2
\end{equation}
where $s'$ and $r$ are the (random) next state and reward from state $s$.
\end{thm}

\begin{dem}
The proof is strictly identical, replacing $V$ with $U$, discarding
all $(1-\gamma)$ terms, and using that $U$ satisfies the centered Bellman
equation \eqref{eq:cBell}. In particular,
the Dirichlet norm is insensitive to adding constants, so the centering
of rewards does not affect the result.
\end{dem}

\paragraph{Advantage function over states, and Dirichlet norm.} Take
$\gamma=1$.
Given a transition $s\to s'$, define
the advantage of $s'$ at $s$ to be 
\begin{equation}
\label{eq:adv}
A(s'|s)\deq \E[r(s,s')]+
U(s')-U(s)
\end{equation}
and likewise the approximate advantage
$A_\theta(s'|s)\deq \E[r(s,s')]+U_\theta(s')-U_\theta(s)$. This is the
``state advantage function'', defined as a function of the next state
$s'$, as opposed to the usual advantage function which is defined on
actions. Once more, this is relevant mostly in a navigation setting where
actions directly correspond to choosing the next state.

Then the Dirichlet norm of $U_\theta-U$ is the average square error of
the advantage function:
\begin{equation}
\label{eq:diradv}
\E_{s\sim \mu}\E_{s'\sim
P(s,s')} (A(s'|s)-A_\theta(s'|s))^2=2\transnorm{U-U_\theta}^2
\end{equation}
by direct substitution.
Therefore, Theorem~\ref{thm:revrelTD} can be restated using this
advantage function.

\begin{cor}
For $\gamma=1$ and for reversible policies, centered approximate TD
is a gradient descent of the average square error $\E_{s\sim \mu}\E_{s'\sim
P(s,s')} (A(s'|s)-A_\theta(s'|s))^2$ of the state advantage function.
\end{cor}

However, for $\gamma<1$ this correspondence breaks down. Indeed, defining
the state advantage function for $\gamma<1$ as
\begin{equation}
A(s'|s)\deq
\E[r(s,s')]+\gamma V(s')-V(s)
\end{equation}
and likewise for $A_\theta$, one checks that
\begin{equation}
\E_{s\sim \mu}\E_{s'\sim
P(s,s')}
(A(s'|s)-A_\theta(s'|s))^2=2\gamma\transnorm{V-V_\theta}^2+(1-\gamma)^2
\munorm{V-V_\theta}^2
\end{equation}
which is not quite the mixed norm minimized by TD: the
weights between the two norms are different.

\section{The Dirichlet Norm and Policy Gradient Bias}

We have proved that with reversible policies, TD approximates the value
function in the Dirichlet norm. This clarifies the behavior of TD for
policy evaluation, but does this help with policy improvement?

Classical results state that if an approximate value function
is $\eps$-close to the true value function (in sup norm), then greedy
policies based on the approximate value function will have cumulated
rewards that are $2\eps/(1-\gamma)$-close to the optimal cumulated
rewards \cite[Prop 2.3.3]{bertsekasvol2_ed4}.

The Dirichlet norm, on the other hand, controls how close policy gradient based
on the true or approximate value functions are to each other: this is
Proposition~\ref{prop:dirgrad} below. Interestingly, this directly holds
with $\gamma=1$, without 
factors $1/(1-\gamma)$.

% TODO: what is policy gradient for $\gamma<1$? For simplicity, we only state the proposition in the case $\gamma=1$
% (expected rewards). TODO: similar result with mixed norm for $\gamma<1$?
% In that case, policy gradient is given by the gradient of the expected
% reward with respect to the parameters of the policy.

% Moreover, for simplicity we assume that we are in a navigation problem,
% i.e., the policy directly controls the probability to jump at the next
% state $s'$. (Otherwise: replace $V(s)$ with $Q(s,a)$, and Dirichlet norms
% on $Q$...)

In a non-episodic setting, policy gradient is defined
as the
gradient of the expected reward under the stationary distribution of the
policy \cite[\S7.4]{bertsekasvol2_ed4}: the goal is to maximize the average reward collected along an
infinitely long trajectory of this policy.

So let $\pi_\pp$ be a policy smoothly parameterized by $\pp$. Let
$\mu_\pp$ be the stationary distribution of $\pi_\pp$. Here we do \emph{not} assume that
policies are reversible.

% Let
% \begin{equation}
% P_\pp(s,a,s')\deq \pi(s,a)\Penv((s,a),s')
% \end{equation}
% be the probability to select action $a$ and end up in state $s'$ when in
% state $s$.
% % Let $P_\pp(s,s')$
% % be the probability to jump from state $s$ to $s'$ when using the policy
% % with parameter $\pp$. $P_\pp$ depends both on the policy and the
% % environment, via \eqref{eq:transprob}. 

The expected reward of the policy with
parameter $\pp$ is
\begin{equation}
\AR(\pp)\deq \sum_{s} \mu_{\pp}(s) R(s)
\end{equation}
with $R(s)$ the expected instantaneous reward in state $s$ (which itself
depends on $\phi$ via the expectation in \eqref{eq:R}). The direction of the policy
gradient update is $\partial_\pp \AR(\pp)$.

The classical policy gradient
theorem \cite[\S7.4.1]{bertsekasvol2_ed4} provides a way to compute
this gradient: it is  an
expectation under the stationary distribution, of the correlation between
expected rewards and action probabilities. \footnote{This assumes the
environment is independent from the parameter $\phi$ used by the agent.} The direction of the
gradient can be expressed
as \cite[Eq.~(7.120)]{bertsekasvol2_ed4} \footnote{Eq.~(7.120) in \cite{bertsekasvol2_ed4} uses centered rewards in the
definition of $\tilde Q$. This is indifferent: since $\sum_a \partial_\pp
\ln \pi_\pp(s,a)=0$, any constant or baseline can be subtracted.}
\begin{align}
\Delta \pp& \deq \partial_\pp \AR(\pp)\nonumber\\&=
\E_{s\sim \mu_\pp,\,a\sim \pi_\pp(s,a),\,s'\sim \Penv((s,a),s')}
\left[ \left(r(s,a,s')+U(s')\right)\partial_\pp \ln \pi_\pp(s,a)\right]
% \E_{s\sim \mu_\pp} \E_{s'\sim P_\pp(s,s')}
% \left[ \left(r(s,s')+U(s')\right)\partial_\pp \ln P_\pp(s,s')\right]
\label{eq:policygrad}
\end{align}
with $U$ the relative value function of the current policy, and $r(s,a,s')$
the random reward incurred during the transition $s\to s'$.

The policy gradient ${\Delta \pp}$ is an expectation over transitions
$(s,a,s')$ from the current policy. As such, an algorithm averaging over
long trajectories from this policy would be a stochastic gradient descent
with expected step $\Delta \pp$. (See also the note after
Theorem~\ref{thm:revTD}.)

Using an approximation of $U$ in \eqref{eq:policygrad}
would result in a bias; we show that this bias is controlled by the
Dirichlet norm of the approximation of $U$.

\begin{prop}
\label{prop:dirgrad}
Let $\hat U$ be any approximation of the relative value function $U$ of the policy
$\pi_\pp$ (undiscounted, $\gamma=1$). Let $\widehat{\Delta \psi}$ be the approximate policy gradient
computed from $\hat U$, namely
\begin{equation}
\widehat{\Delta \pp}\deq 
\E_{s\sim \mu_\pp,\,a\sim \pi_\pp(s,a),\,s'\sim \Penv((s,a),s')}
\left[ \left(r(s,a,s')+\hat U(s')\right)\partial_\pp \ln \pi_\pp(s,a)\right]
%\E_{s\sim \mu_\pp}\E_{s'\sim P_\pp(s,s')}\left[ \left(r(s,s')+\hat U(s')\right)\partial_\pp \ln P_\pp(s,s')\right]
\end{equation}
Then the bias of this approximate policy gradient is at most
\begin{equation}
\norm{\widehat{\Delta \pp}-\Delta\pp}^2\leq
2\transnorm{U-\hat U}^2 \cdot \left(
\E_{s\sim \mu_\pp}\E_{a \sim \pi_\pp(s,a)} \norm{\partial_\pp \ln
\pi_\pp(s,a)}^2
\right)
% =
% \transnorm{V-\hat V}^2 \cdot \left(
% \E_{s\sim \mu_\pp}\E_{a\sim \pi_\pp(a|s)} \norm{\partial_\pp \ln
% \pi_\pp(a|s)}^2
% \right)
\end{equation}
\end{prop}

As a consequence, if $U$ tends to $\hat U$ in Dirichlet norm then the
bias tends to $0$. Of course this is the bias over a single step of
policy gradient. \cite{SLRB2011_bias} contains a full study of the
asymptotic bias produced by a bias at each step of a stochastic gradient
descent, under convexity assumptions (which would hold close to a
nondegenerate local minimum in typical cases); in particular, under
strong convexity, a bounded bias at each step of a gradient descent only
produces a bounded deviation from the true trajectory
\cite[Prop.~3]{SLRB2011_bias}.

Since $\transnorm{\cdot}\leq \munorm{\cdot}$, the inequality also holds
with $\munorm{U-\hat U}$, but is less sharp (for instance,
$\transnorm{\cdot}$ is
insensitive to adding a constant to $\hat U$), sometimes much less so
depending on the spectral gap $\beta$ in \eqref{eq:eqnorms}.

The last factor, $\E_{s\sim \mu_\pp}\E_{a\sim \pi_\pp(s,a)}
\norm{\partial_\pp \ln \pi_\pp(s,a)}^2$, does not depend on the way the
value function is approximated: it depends only on 
the way policies are parameterized. It is equal to the
trace of the Fisher information matrix of the policy $\pi_\pp(s,a)$ with respect to $\pp$. Thus,
there is a clear contribution from value function approximation, and
another from the geometry of the space of policies.

% The quantity $P_\pp(s,s')$ may not be readily available, except in
% a navigation problem when the policy directly sets these transition
% probabilities, or when an exact model of the environment is available.
% Thus, once more, this is mostly relevant for navigation problems only.

% TODO: this assumes that the transition matrix of the environment does not
% depend on the parameter of the policy. In physical systems or in
% adversarial games this might be false if the action
% parameters of an agent can somehow be measured by the environment.
% If the parameter $\psi$ can influence the environment then the correct
% inequality is
% \begin{equation}
% \norm{\widehat{\Delta \pp}-\Delta\pp}^2\leq
% \transnorm{V-\hat V}^2 \cdot \left(
% \E_{s\sim \mu_\pp}\E_{s'\sim P_\pp(s,s')} \norm{\partial_\pp \ln
% P_\pp(s,s')}^2
% \right)
% \end{equation}

\begin{dem}
The proof is essentially the Cauchy--Schwarz inequality after subtracting
a suitable baseline.

For short,
denote
\begin{equation}
\xi(s,a,s')\deq \mu_\pp(s)\pi_\pp(s,a)\Penv((s,a),s')
\end{equation}
the stationary distribution over transitions $(s,a,s')$ when using policy
$\pi_\pp$.

Gradients of log-probabilities have expectation $0$, so for any
state $s$,
\begin{equation}
\E_{a\sim \pi_\pp(s,a)}\, \partial_\pp \ln \pi_\pp(s,a)=0
\end{equation}
therefore, in the policy gradient formula \eqref{eq:policygrad} we can
subtract any baseline depending on $s$, as is often done in practice:
\begin{equation}
\Delta \pp =
\E_{(s,a,s')\sim \xi}
\left[ \left(r(s,a,s')+U(s')-U(s)\right)\partial_\pp \ln \pi_\pp(s,a)\right]
\end{equation}
and likewise
\begin{equation}
\widehat{\Delta\pp}=
\E_{(s,a,s')\sim \xi}
\left[ \left(r(s,a,s')+\hat U(s')-\hat U(s)\right)\partial_\pp \ln \pi_\pp(s,a)\right]
\end{equation}
therefore
\begin{equation}
\widehat{\Delta\pp}-\Delta\pp=\E_{(s,a,s')\sim \xi}
\left[ \left(\hat U(s')-U(s')-\hat U(s)+U(s)\right)\partial_\pp \ln \pi_\pp(s,a)\right]
\end{equation}
so by the Cauchy--Schwarz inequality
\begin{multline}
\norm{\widehat{\Delta\pp}-\Delta\pp}^2
\leq \left(\E_{(s,a,s')\sim \xi}\left(\hat U(s')-U(s')-\hat
U(s)+U(s)\right)^2\right)\cdot\\
\left(\E_{(s,a,s')\sim \xi} \norm{\partial_\pp \ln
\pi_\pp(s,a)}^2\right)
\end{multline}

Now marginalizing $\xi(s,a,s')$ over $a$ yields $\mu_\pp(s)P_\pp(s,s')$
by definition \eqref{eq:transprob}.
So by definition of the Dirichlet norm \eqref{eq:dir}, the first factor
above is exactly $2\transnorm{\hat U-U}^2$.
\end{dem}

Note that the Dirichlet norm itself depends on the parameter $\phi$, via
$P$ and $\mu$.

\section{Discussion and Conclusion}

\paragraph{Advantage function, and Dirichlet norm versus $L^2$ norm.} Converging to the true
value function in $L^2$ norm emphasizes getting the correct value at each
state. On the other hand, converging to the true value function in
Dirichlet norm emphasizes getting the correct \emph{differences of
values} between consecutive states: this is clear from the
definition \eqref{eq:dir}. Getting these differences right amounts to
being able to compare the values of states.
Proposition~\ref{prop:dirgrad} formalizes this intuition: the smaller the
error in Dirichlet norm, the smaller the bias in policy gradient.
This is also directly related to the
advantage function (Eq.~\ref{eq:diradv}) for $\gamma=1$:
% Given a transition $s\to s'$, define
% the advantage of $s'$ at $s$ to be $A(s'|s)\deq
% U(s')+\E[r(s,s')]-U(s)$ and likewise the approximate advantage
% $A_\theta(s'|s)\deq U_\theta(s')+\E[r(s,s')]-U_\theta(s)$. Then
% the Dirichlet norm of $U_\theta-U$ is the average square error of the
% advantage function:
% \begin{equation}
% \E_{s\sim \mu}\E_{s'\sim P(s,s')}
% (A(s'|s)-A_\theta(s'|s))^2=2\transnorm{U-U_\theta}^2
% \end{equation}
for reversible policies and $\gamma=1$, TD is just a gradient descent of the $L^2$
error of the state advantage function \eqref{eq:adv}. Here, advantages are
computed on
next states $s'$, rather than on actions as is more common; so once more this is mostly
relevant for navigation problems or when a good model of the environment
is available.

\paragraph{On convergence speed when $\gamma\to 1$.} In
Theorem~\ref{thm:revTD}, the properties of TD do not deteriorate when
$\gamma\to 1$: the Dirichlet norm $\transnorm{V_\theta-V}$ will decrease
at a rate that does not depend on $\gamma$. More precisely, TD with step
size $\eta$ is a gradient descent of the mixed norm with step size
$\eta/2$. On the other hand, the convergence proof from
\cite{tsitsiklis1997approxtd} in the linear case provides smaller and
smaller learning rates when $\gamma\to 1$: the rate of decrease of the
error $\theta-\theta^*$ is given by Lemma 9 from
\cite{tsitsiklis1997approxtd}, which contains a $(\gamma-1)$ factor
(though this can be improved using the spectral gap of the Markov chain).

\paragraph{Optimizing among reversible policies.} The reversibility
constraint on the policy is obviously a major restriction of these
results. \label{sec:revpolicies}

Still, the space of reversible
policies is quite large. For instance, for any positive function $f$ on
any unoriented graph, the celebrated Metropolis--Hastings construction
\cite[\S 3.2.2]{LPW2009_markov}
provides a reversible random walk on the edges of the graph, whose
stationary distribution is proportional to $f$. The
probability to jump from $s$ to $s'$ is set to
\begin{equation}
P_f(s,s')=\min\left(
\frac{1}{\deg(s)},\frac{f(s')}{f(s)\deg(s')}
\right)
% P_f(s,s')=\begin{cases}
% \frac{1}{\deg(s)} & \text{if } f(s')\geq f(s)
% \\
% \frac{1}{\deg(s)} \,\frac{f(s')}{f(s)} & \text{if } f(s')<f(s)
% \end{cases}
\end{equation}
for any adjacent states $s\neq s'$ in the graph. Then
$\mu(s)\deq \frac{f(s)}{\sum_{s'} f(s')}$ is a reversible stationary
distribution of $P_f$. More generally, if $P_0$ is any ``default'' Markov
chain, then the Markov chain
$P_f(s,s')=\min\left(P_0(s,s'),\frac{f(s')P_0(s',s)}{f(s)}\right)$ for
$s\neq s'$, is
reversible with respect to this same $\mu$.

Thus, in some cases it would be possible to explicitly keep the policy in
a space of reversible policies.
In particular, any navigation problem where the agent directly selects
the next state among a set of neighbors of the current state, defines an
unoriented graph. For such problems,
policies targeting any stationary distribution $f$ over
states can be obtained by parameterizing a family of positive functions
$f$ over the state space in any convenient way, and setting the family
of policies to the Metropolis--Hastings Markov chain $P_f$ for $f$ in
this family. A natural candidate would be Gibbs distributions of
the form $f(s)=\exp (\beta V_\theta(s))$ where $V_\theta$ is the family
used to approximate the value function: for large $\beta$ this targets
high-value states.

\paragraph{Conclusion.} We have proved that the unmodified approximate TD algorithm is exactly a gradient descent of
the Dirichlet norm of the error between the true and approximate value
functions, provided the policy is reversible. 
The Dirichlet norm also controls the bias of approximate policy gradient
and the $L^2$ error on the advantage function over states,
even for non-reversible policies. However, the reversibility condition is
restrictive: only for navigation problems can one easily maintain the
policy within a set of reversible policies.

Thus, at least for navigation problems, the Dirichlet norm provides a
coherent theoretical picture of what approximate TD does.

\bibliographystyle{alpha}
\bibliography{tdconv}

\end{document}